\documentclass[runningheads]{llncs}

\usepackage[T1]{fontenc}
\usepackage{graphicx,verbatim}
\usepackage{hyperref}
\usepackage{footmisc}
\usepackage{amssymb}
\usepackage{amsmath}
\usepackage{mathtools}
\usepackage{upgreek}
\usepackage{subcaption}
\usepackage{orcidlink}

\DeclareMathAlphabet\mathbfcal{OMS}{cmsy}{b}{n}

\begin{document}
\newcommand{\bh}{\mathbf{h}}
\newcommand{\bx}{\mathbf{x}}
\newcommand{\by}{\mathbf{y}}
\newcommand{\bu}{\mathbf{u}}
\newcommand{\bv}{\mathbf{v}}
\newcommand{\bz}{\mathbf{z}}
\newcommand{\un}{\mathds{1}}
\newcommand{\calC}{\mathcal{C}}
\newcommand{\calD}{\mathcal{D}}
\newcommand{\calL}{\mathcal{L}}
\newcommand{\calN}{\mathcal{N}}
\newcommand{\calP}{\mathcal{P}}
\newcommand{\calT}{\mathcal{T}}

\title{Pix2Rep-v2: Data-Efficient Representation Learning for Dense Medical Imaging Applications}
\titlerunning{Pix2Rep-v2: Data-Efficient Dense Representation Learning}
\author{Sofiane Sifaoui\inst{1}\orcidlink{0009-0005-1530-3092} \and
Elsa Angelini\inst{1}\orcidlink{0000-0002-1602-300X} \and
Solenn Toupin\inst{2,3}\orcidlink{0000-0001-9967-0044} \and
Théo Pezel\inst{2,3}\orcidlink{0000-0002-6716-8444} \and
Loïc Le Folgoc\inst{1}\orcidlink{0000-0002-5156-6616}}
\authorrunning{S. Sifaoui et al.}
\institute{LTCI, Télécom Paris, Institut Polytechnique de Paris, Palaiseau, France \and
MIRACL.ai Laboratory, Hôpital Universitaire Lariboisière (AP-HP), Paris, France \and Université Paris Cité, Inserm MASCOT, Paris, France \\
\email{sofiane.sifaoui@ip-paris.fr}}
\maketitle
\begin{abstract}

Dense self-supervised learning (SSL) is a powerful paradigm for learning without annotations the local descriptors required to solve dense medical imaging tasks. We present \texttt{Pix2Rep-v2}, a framework for SSL of pixel- and voxel-level representations suitable for few-shot downstream applications. \texttt{Pix2Rep-v2} addresses the main challenges of dense SSL by leveraging a redundancy reduction objective at the pixel-level with a principle of equivariance of dense representations, that scales efficiently to 3D or wide field-of-view applications. We evaluate our method on four datasets, across multiple tasks, multiple modalities and anatomical structures using multiple backbones in 2D and 3D, and under various data regimes. As an alternative to linear probing or full fine-tuning on the downstream task, we also propose an in-context variant, without downstream training, based on a dense prototype approach. \texttt{Pix2Rep-v2} shows substantially higher data-efficiency in few-shot scenarios compared to fully supervised baselines, and is competitive with the state-of-the-art \textit{e.g.,} $+9.3$ Dice points in one-shot segmentation on the M\&Ms-2 dataset. Our code and pre-trained models are publicly available at \url{https://github.com/BioMedTP/pix2rep-v2}.

\end{abstract}
\section{Introduction}

Supervised deep learning has considerably advanced automation of dense medical imaging tasks such as segmentation \cite{ronneberger_olafand_fischer_u-net_2015,bernard_deep_2018,hatamizadeh_swin_2022}. However scarcity of pixel-level annotations remains a bottleneck for the development of new AI solutions on new applications, on different modalities or for deployment data from new scanners. 

Several paradigms have emerged to circumvent this bottleneck, starting with transfer learning or domain generalization~\cite{9961940} that transport pre-existing models to the target task or dataset, and semi-supervised learning~\cite{tran_manueland_wagner_s5cl_2022}, which leverages unlabeled data. Recently, generalist~\cite{simeoni_dinov3_2025} or specialized~\cite{wasserthal_totalsegmentator_2023} foundation models trained from massive, diverse datasets promise to solve a broad range of applications in zero-shot or by fine-tuning on the application of interest. Such models demand either thousands of densely annotated scans~\cite{wasserthal_totalsegmentator_2023} or, preferably, effective self-supervised pre-training recipes to train at scale~\cite{simeoni_dinov3_2025,jacob_towards_2025,fu2026development} on up to millions of unlabeled scans. As a practical alternative, practitioners also look for effective solutions to train small task-specific and data-specific models from scratch on premise at minimal annotation cost.

We present an SSL framework for both purposes, \texttt{Pix2Rep-v2}, that addresses several challenges of dense contrastive learning methods~\cite{o.pinheiroUnsupervisedLearningDense2020,goncharov_mikhailand_soboleva_vox2vec_2023,seince_dense_2024}. Existing approaches~\cite{o.pinheiroUnsupervisedLearningDense2020,goncharov_mikhailand_soboleva_vox2vec_2023} contrast pixel-level representations from overlapping regions of two crops. To preserve sufficient overlap between views, they adopt milder spatial augmentation strategies that ultimately limit performance. Secondly, contrasting individual local feature vectors incurs high computational and memory costs due to the large number of negatives. We overcome these challenges through an equivariance-based formulation that relies on a single arbitrary spatial augmentation, and on a non-contrastive redundancy reduction formulation. \texttt{Pix2Rep-v2} also returns high-quality pixel-level representations with strong local semantics straight out of pre-training, enabling a direct in-context downstream use (no fine-tuning). In summary, we make the following contributions:

\begin{itemize}
    \item[$\bullet$] We present $\texttt{Pix2Rep-v2}$, addressing challenges of SOTA dense SSL approaches via an efficient redundancy reduction objective at the pixel-level and an aggressive multiscale approach;
    \item[$\bullet$] We evaluate $\texttt{Pix2Rep-v2}$ on multiple datasets, across multiple tasks (segmentation, video propagation), multiple modalities (cine MRI, CT), multiple anatomical structures (cardiac, abdominal), multiple data regimes (one-shot, few-shots, many-shots), multiple backbones, in 2D and 3D; 
    \item[$\bullet$] To better investigate the intrinsic quality of $\texttt{Pix2Rep-v2}$'s pixel-level representations, we propose in addition to the fully fine-tuned and linear probing variants, a parameter-free training-free in-context version.
\end{itemize}

\section{Related Work}

\indent\textbf{Self-Supervised Learning} first emerged as a paradigm to learn global image-level representations, based on various objectives: contrastive losses~\cite{chen_simple_2020}, redundancy reduction~\cite{zbontar_barlow_2021}, self-distillation~\cite{simeoni_dinov3_2025}, masked image modeling (MIM)~\cite{he_masked_2022}. Dense SSL instead aims to learn pixel-level or patch-level representations suitable for dense downstream tasks. Most dense SSL methods adapt contrastive~\cite{goncharov_mikhailand_soboleva_vox2vec_2023,kats_self-supervised_2024,seince_dense_2024} or MIM objectives, or joint-embeddings~\cite{simeoni_dinov3_2025}, except \texttt{BT-UNet}~\cite{punn_bt-unet_2022}, which is based on redundancy reduction, but pre-trains only the encoder of the U-Net~\cite{ronneberger_olafand_fischer_u-net_2015}. 

Pixel-level contrastive methods~\cite{o.pinheiroUnsupervisedLearningDense2020,goncharov_mikhailand_soboleva_vox2vec_2023,kats_self-supervised_2024} typically aim to align representations of the same anatomical points from two partially overlapping image crops. In addition, contrastive methods (incl.~\texttt{Pix2Rep}~\cite{seince_dense_2024}) sparsely sample the scans to avoid an explosion of the negative sample size. 3D applications present computational challenges for these methods that~\cite{goncharov_mikhailand_soboleva_vox2vec_2023,kats_self-supervised_2024} solve via a dedicated 3D Feature Pyramid Network (FPN) coarse-to-fine representation. Alternatively, patch-level SSL methods~\cite{he_masked_2022,simeoni_dinov3_2025,jacob_towards_2025,fu2026development} train the encoder only, whereas the backbone decoder is trained from scratch during downstream fine-tuning, potentially reducing few-shot performance. 

\textbf{Foundation Models.} SSL recipes can be deployed at scale to train generalist foundation models such as \texttt{DINOv3}~\cite{simeoni_dinov3_2025} or specialized foundation models \textit{e.g.,} \texttt{CineMA}~\cite{fu2026development} and \cite{jacob_towards_2025} for cardiac MRI applications. \texttt{MAE}~\cite{he_masked_2022} or \texttt{DINO}-based~\cite{oquab2024dinov} pre-training is standard for such models. Alternatively, other foundation models such as \texttt{TotalSegmentator}~\cite{wasserthal_totalsegmentator_2023} are trained with label supervision. Lastly, \texttt{SAM 3}~\cite{carion2026sam} or in the medical domain \texttt{MedSAM2}~\cite{MedSAM2} enable segmentation or video propagation with guidance from various prompts: points, boxes or segmentation masks.

\textbf{In-Context Learning} allows to solve new segmentation tasks in few-shots through example-guided inference with no task-specific fine-tuning. In particular, \texttt{ALPNet}~\cite{ouyang_self-supervised_2022} adopts a prototype-based approach, which \texttt{ProtoSAM}~\cite{ayzenberg_protosam_2025} extends by leveraging \texttt{DINOv2}~\cite{oquab2024dinov} and \texttt{SAM}~\cite{kirillovSegmentAnything2023a} capabilities. In this paper, we also propose a straightforward, scalable alternative based on a dense prototype approach.

\section{Methods}
\label{sec: Methods}

\begin{figure}[t]
\includegraphics[width=\textwidth]{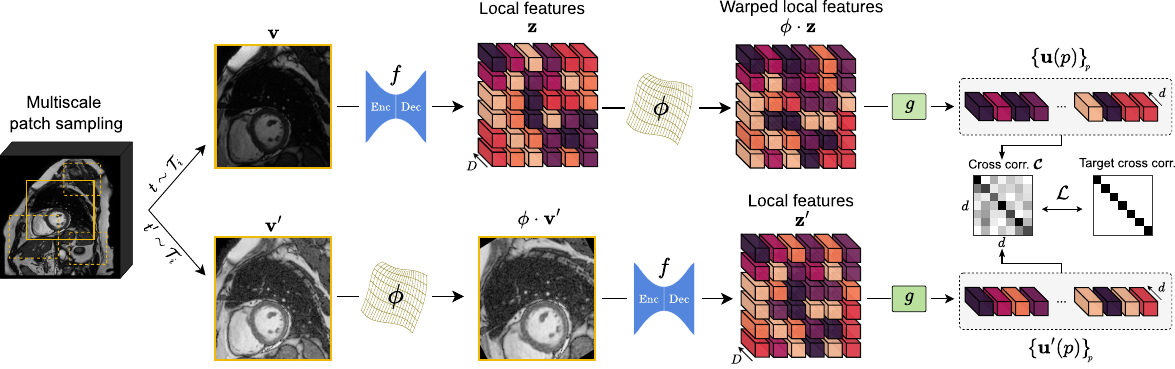}
\caption{\textbf{Overview of {Pix2Rep-v2}}.
Multi-scale patches are sampled from input images and transformed by two intensity augmentations producing two views: $\bv$ and $\bv'$. A random spatial transformation $\phi$ is applied to $\bv'$ to map to a new viewpoint, then pixel representations $\bz'$ are extracted using an encoder–decoder backbone $f$. Asymmetrically, $\bv$ is processed by $f$ before its pixel representations $\bz$ are mapped to the new viewpoint $\bv$ by action of $\phi$. The resulting paired pixel-level representations are fed to an MLP projector $g$, from which an empirical cross-correlation matrix is computed. The training loss encourages equivariance of pixel representations while reducing feature redundancy.}
\label{fig: Pipeline}
\end{figure}

\texttt{Pix2Rep-v2} extends the \texttt{Pix2Rep} dense SSL paradigm, with a redundancy reduction (vs.~contrastive) loss function, an aggressive multiscale patch sampling strategy, 3D support and in-context capabilities.

We pre-train an arbitrary encoder-decoder backbone $f:\mathbb{R}^{H\times W\times C}\rightarrow \mathbb{R}^{H\times W\times D}$ that maps from pixel-space to representation-space, using an unlabeled dataset $\mathcal{D}\triangleq \{\bx\in \mathbb{R}^{H\times W\times C}\}$ of image patches, by enforcing two constraints on the representations: (c.1) invariance to photometric augmentations and equivariance to spatial transformations; and (c.2) informativeness and non-redundancy component-wise. An MLP projection head $g: \mathbb{R}^{H\times W\times D}\rightarrow \mathbb{R}^{H\times W\times d}$ maps pixel representations to the space where the redundancy reduction loss is computed.

For a given image patch $\bx$, we generate two random photometric transformations $t,t'\sim \calT_i$, and one random spatial transformation $\phi\sim \mathcal{T}_s$, which maps to a new viewpoint. Photometric augmentations include random bias fields, Gamma distortions, blur, intensity rescaling, Gaussian noise, and intensity inversion. Spatial transformations include random flips, rotations, zooms, and B-spline elastic deformations. 
Applying $t,t'$ to $\bx$ yields two views $\bv\triangleq t(\bx), \bv'\triangleq t'(\bx)$. Then, asymmetrically: we transport $\bv'$ to the new viewpoint by action of $\phi$ on $\bv'$ \textit{i.e.}, $\phi\cdot \bv'\triangleq \bv'\circ\phi^{-1}$ then compute its pixel representations $\bz'\triangleq f(\phi\cdot \bv')$ from the new viewpoint; whereas for $\bv$, we compute pixel representations $\bz=f(\bv)$ from the initial viewpoint, then transport $\bz$ to the new viewpoint: $\phi \cdot \bz = \phi \cdot f(\bv)$.

Finally, we consider all paired representations $\{(\phi\cdot\bz)(p), \bz'(p)\}$, across all $P$ pixels in all image patches in a batch, which we project through $g(\cdot)$ then normalize to zero mean, unit standard deviation, yielding paired vectors $\{\bu(p), \bu'(p)\}$. 

Let $\mathbfcal{C}$ be the cross-correlation matrix with coefficient $\mathbfcal{C}_{ij}\triangleq \sum_{p} \bu(p)_i \bu'(p)_j$, where $1\leq i,j\leq d$ index two components of the projected representations. 
Computing and storing $\mathbfcal{C}\in \mathbb{R}^{d\times d}$ on GPU is straightforward, unlike the similarity matrix in contrastive approaches~\cite{seince_dense_2024,goncharov_mikhailand_soboleva_vox2vec_2023,kats_self-supervised_2024}, which typically scales with the square of the number of pixels $P\gg d$.
We minimize the redundancy reduction loss $\calL$ of Eq.~\eqref{eq: Barlow Twins loss}, defined as in Barlow Twins~\cite{zbontar_barlow_2021}:
\begin{equation}
    \calL \triangleq \sum_{i\leq d} \left(\mathbfcal{C}_{ii}-1\right)^2 
    + \lambda \sum_{i\leq d}\sum_{j\neq i} \mathbfcal{C}_{ij}^2
    \label{eq: Barlow Twins loss}
\end{equation}

\noindent\textbf{Multiscale Patch Sampling.} Each batch $\{\bx\in \mathbb{R}^{H\times W\times C}\}$ contains image patches (typically $H\coloneqq W \coloneqq 128$) extracted from whole scans. We extract on-the-fly one random patch of random dimensions $H_0\times W_0$ per scan, and resize it to $H\times W$ without changing aspect ratio. $H_0\coloneqq W_0\coloneqq \alpha \cdot \text{min}(H_{\text{scan}},W_{\text{scan}})$ is $\alpha$ times the smallest dimension (width or height) of the whole scan, where $\alpha \sim \mathcal{U}(\alpha_{\text{min}},\alpha_{\text{max}})$ is uniformly sampled at random (typically $\alpha_{\text{min}}\coloneqq 0.33$ and $\alpha_{\text{max}}\coloneqq 0.75$). 
This exposes the pre-trained model to a large variety of patches and teaches it to deal with input images at multiple scales (Fig.~\ref{fig: Pipeline}).\\

\noindent\textbf{Downstream Segmentation.} We train a segmentation head (Linear + Softmax) on top of the backbone $f(\cdot)$, discarding $g(\cdot)$. In linear probing, the backbone is frozen; in fine-tuning, we fine-tune the whole model. Either way, different from pre-training, during task-specific training we resample all scans to a fixed spacing before extracting $H\times W$ image patches; we proceed identically at inference time.\\

\noindent\textbf{3D Backbone.} \texttt{Pix2Rep}~\cite{seince_dense_2024} shows that downstream performance benefits substantially from large $D$ values. However storing explicitly many full-resolution feature maps, as output by the backbone decoder, is prohibitive memory-wise for 3D applications. We propose instead an implicit 3D U-Net backbone inspired by~\cite{marimont2022implicit}, where the upper blocks of the decoder are replaced by MLP layers (we refer the reader to the \href{https://github.com/BioMedTP/pix2rep-v2}{code} for details). In this implicit U-Net, the output representations can be computed for a smaller specified set of point coordinates rather than on the regular pixel grid. We randomly sample $2^{17}$ ($>10^5)$ coordinates per 3D patch, on which to evaluate Eq.~\eqref{eq: Barlow Twins loss}. This is more than $100\times$ the number of points usually sampled in contrastive dense SSL~\cite{goncharov_mikhailand_soboleva_vox2vec_2023,kats_self-supervised_2024} (1024 points per volume). Different from the coarse-to-fine representations extracted from the 3D FPN backbone of~\cite{goncharov_mikhailand_soboleva_vox2vec_2023,kats_self-supervised_2024}, the implicit 3D U-Net backbone implicitly extracts a large number of features at high-resolution.\\

\noindent\textbf{In-Context Segmentation.} Given a backbone pre-trained with \texttt{Pix2Rep-v2} and a support set $X_S=\{(\bx^{(s)},\by^{(s)})\}_{s\in S}$ of images with their corresponding ground truth (GT) segmentations, we wish to predict segmentation maps for all images in a query set $X_Q=\{\bx^{(q)}\}_{q\in Q}$ without any task-specific fine-tuning.

We adopt a dense prototype approach whereby the \texttt{Pix2Rep-v2} representations of all pixels in all images of $X_S$ are gathered to form the prototype set $\calP$. Then, for any given pixel in a query image, we compute its \texttt{Pix2Rep-v2} (projected) representation
$g(\bz(p))$, perform a nearest neighbor search in $\calP$ w.r.t.~cosine similarity, and assign the label of this support pixel to the query pixel. This type of nearest neighbor search on large sets (up to $10^9$ elements) of high-dimensional vectors can be performed extremely efficiently with the \texttt{FAISS}~\cite{faiss_douze} library, yielding a straightforward, parameter-free and scalable strategy.\\

\noindent\textbf{Zero-Shot 3D+t Video Propagation.} We consider $3D+t$ applications where for each patient, the GT segmentation is available on a reference frame, and we wish to propagate it to the rest of the time series, without label-specific fine-tuning, using the pre-trained \texttt{Pix2Rep-v2} representations. 

We adopt a propagation strategy across consecutive frames $t-1 \rightarrow t$ via a dense prototype approach. 
For any frame $t$ in the series, the prototype set includes $\calP_{t-1}$, the \texttt{Pix2Rep-v2} representations of all pixels in the previously segmented frame $t-1$. To reduce error accumulation over several frames, we add to the prototype set the representations of all pixels extracted from an ``anchor'' frame, here the reference frame $t_0$ where the GT is available \textit{i.e.,} $\calP_{t-1}\cup\calP_{t_0}$.
We segment the frame $t$ by assigning to any given pixel, with representation $g(\bz(p))$, the label of its nearest neighbor in the prototype set w.r.t.~cosine similarity.

\section{Experiments and Results}

We evaluate the \texttt{Pix2Rep-v2} framework on cardiac MRI segmentation and video propagation, as well as on abdominal CT multi-organ segmentation.\\

\begin{figure}[!t]
\centering
\includegraphics[width=1.0\textwidth]{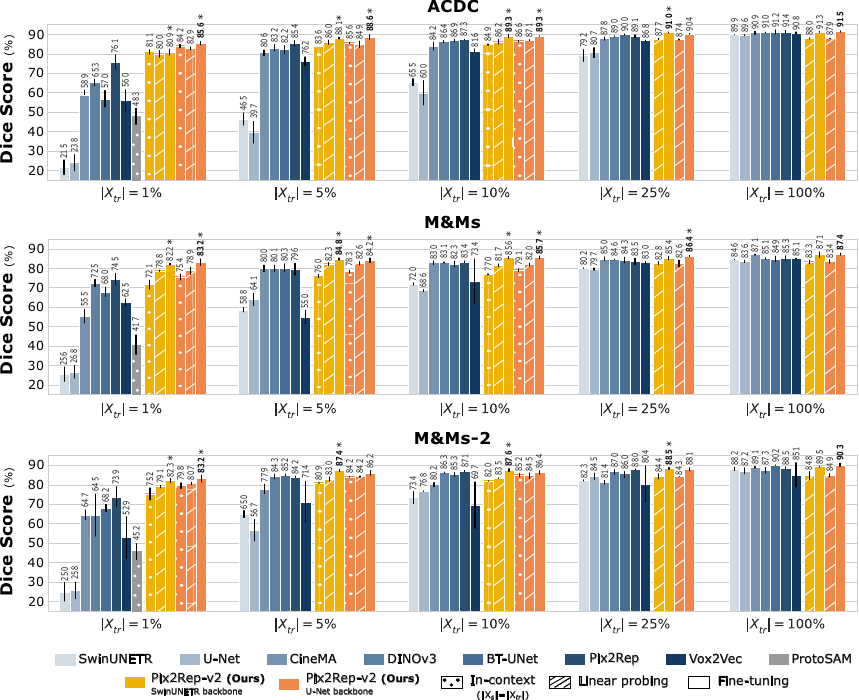}
\caption{\textbf{Cardiac MRI segmentation results}
per cohort: ACDC, M\&Ms and M\&Ms-2. For \texttt{Pix2Rep-v2}: in-context, linear-probing or fine-tuning with either backbone ({U-Net} or {Swin-UNETR}). For comparison: {U-Net}, {Swin-UNETR} baselines trained from scratch, fine-tuned foundation models (\texttt{CineMA}, \texttt{DINOv3}), fine-tuned SSL methods (\texttt{BT-UNet}, \texttt{vox2vec}, \texttt{Pix2Rep}) and in-context \texttt{ProtoSAM}. Colored bar + number $\equiv$ mean Dice over 3 runs (with different seeds and training subjects). Black line $\equiv$ standard deviation. Best Dice indicated in bold. \\ ($\boldsymbol{\ast}$) indicates a statistically significant improvement of fine-tuned \texttt{Pix2Rep-v2} over the best baseline in each data regime (Wilcoxon signed-rank test, $p < 0.05)$.}
\label{fig: MRI segmentation results}
\end{figure}

\begin{figure}[t]
\includegraphics[width=\textwidth]{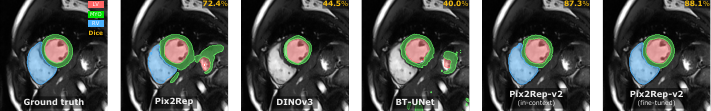}
\caption{\textbf{Qualitative segmentation results} on M\&Ms-2 with $|X_{tr}|=1\%$.}
\label{fig: Qualitative results}
\end{figure}

\noindent\textbf{Implementation Details.} The framework is implemented in \texttt{PyTorch} and publicly available (source code, hyperparameter configurations and pretrained models). We set $\lambda\coloneq 5\cdot 10^{-3}$, $D\coloneqq 1024$, $d\coloneqq 256$ for 2D applications; and $D\coloneqq 256$, $d\coloneqq 128$ for $3D$ applications. For pretraining, we use a learning rate of $5e^{-4}$ in 2D (resp. $1e^{-4}$ in 3D) for the backbone, following a cosine annealing schedule and AdamW optimizer. During finetuning, this base learning rate is divided by a factor 10. We generally pre-train on 4 H100 GPUs for 200 epochs in less than a day, and fine-tune on one V100 GPU for 100 epochs in few hours.\\

\noindent\textbf{Datasets.} \textbf{ACDC}, \textbf{M\&Ms}, \textbf{M\&Ms-2}~\cite{bernard_deep_2018,campello_multi-centre_2021,martin-isla_deep_2023} 
contain 3D short-axis cardiac cine MRI images of 150, 345 and 360 subjects respectively, including GT annotations at End-Systole (ES) and End-Diastole (ED) for the left ventricle, right ventricle and myocardium. We use the provided splits, with $100/209/200$ subjects for training and {$50/136/160$} for testing. These datasets also include the full $3D+t$ cine MRI sequence, which we use in the video propagation downstream task.
\textbf{AMOS}~\cite{ji_amos_2022} includes 3D abdominal CT scans of 500 subjects with multi-organ GT annotations, split between $200$ training, $100$ validation and $200$ test scans. As GT annotations are not disclosed for the original test set, we form a new disjoint split by rearranging subjects: (s.1) 400 for pre-training, including 200 with GT annotations for fine-tuning; (s.2) 100 with GT for testing. Furthermore, $1900$ unlabeled CT scans are also included in the dataset, which we add for pre-training. CT scans are min-max normalized in $[0,1]$, clipping at $HU_{\text{min}}\coloneqq -200$ and $HU_{\text{max}}\coloneqq 300$. In 3D, we extract patches of size 192$\times$192$\times$64.\\

\noindent\textbf{Experimental Setup \& Evaluation.} For pre-training, we use the entirety of the raw training data noted $X_{pre}$, without GT labels. For linear probing or fine-tuning on segmentation tasks, we use a smaller number of training images with their segmentation
labels to simulate one-shot, few-shot, many-shot regimes \textit{e.g.,} $X_{tr}$ is \{1, 5, 10, 25, 100\}\% of the training set. $X_{pre}$ (resp.~$X_{tr}$) is randomly split between training data (90\%) and validation data (10\%) during runs. Test data is only used for the final evaluation. 
For anatomical cardiac MRI applications, we form a single pre-training set $X_{pre}$, using the $3D+t$ raw data in the combined ACDC, M\&Ms and M\&Ms-2 training sets. However, we conduct task-specific fine-tuning separately on each dataset. 

We demonstrate \texttt{Pix2Rep-v2}'s effectiveness with various backbones. For cardiac MRI experiments, we favor 2D backbones due to the large slice thickness, specifically 2D U-Net~\cite{ronneberger_olafand_fischer_u-net_2015} and Swin-UNETR~\cite{hatamizadeh_swin_2022}. Abdominal CT experiments use the implicit 3D U-Net backbone (section~\ref{sec: Methods}).

For video propagation on ACDC, M\&Ms, M\&Ms-2: for each subject, we take for given the GT segmentation at ED and propagate ED$\rightarrow$ES, and vice-versa.

We quantify performance across all applications via the 3D Dice score. We report the 3D Dice averaged over the segmented structures (as well as over ED and ES for ACDC, M\&Ms, M\&Ms-2 datasets), and over the test set.\\

\noindent\textbf{Comparison to the SOTA.} For video propagation, we compare to \texttt{SAM 3}~\cite{carion2026sam} and \texttt{MedSAM2}~\cite{MedSAM2} using the reference frame's GT mask as prompt, as well as to \texttt{Pix2Rep}-based video propagation using their contrastive representations coupled with our proposed propagation mechanism (section~\ref{sec: Methods}). 

For segmentation, a sound baseline to assess the gain in data-efficiency due to \texttt{Pix2Rep-v2} pre-training is to skip pre-training \textit{i.e.,} train the same backbone and segmentation head from scratch. In addition, we compare against the following SOTA methods: for dense SSL, \texttt{vox2vec}~\cite{goncharov_mikhailand_soboleva_vox2vec_2023}, its extension~\cite{kats_self-supervised_2024}; \texttt{Pix2Rep}~\cite{seince_dense_2024} with our proposed multiscale patch sampling but their contrastive loss; for redundancy reduction-based methods, \texttt{BT-UNet}~\cite{punn_bt-unet_2022}; recent foundation models fine-tuned on the tasks, specifically \texttt{DINOv3}~\cite{simeoni_dinov3_2025}, as well as the \texttt{MAE}-based \texttt{CineMA}~\cite{fu2026development} for cardiac applications; for one-shot in-context prototype-based methods, \texttt{ProtoSAM}~\cite{ayzenberg_protosam_2025}. In 3D, we compare with all natively 3D methods in the previous list.\\

\begin{figure}[t]
\centering
\begin{minipage}[t]{0.51\textwidth}
    \centering
    \includegraphics[width=\textwidth]{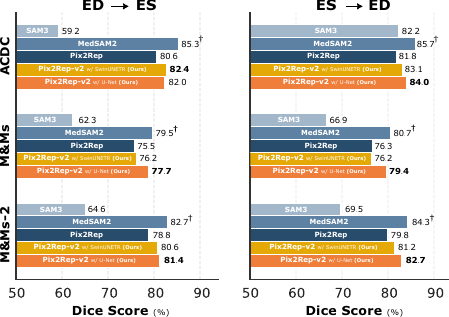}
    \captionof{figure}{\textbf{Cine MRI video propagation results}. Best viewed zoomed-in, in color. \texttt{Pix2Rep-v2}, \texttt{SAM3} and \texttt{Pix2Rep}'s predictions are zero-shot, whereas \texttt{MedSAM2}~($\dagger$) is data contaminated: its training set includes ACDC, M\&Ms and M\&Ms-2 scans and GT annotations.}
    \label{fig: video propagation results}
\end{minipage}
\hfill
\begin{minipage}[t]{0.47\textwidth}
    \centering
    \includegraphics[width=\textwidth]{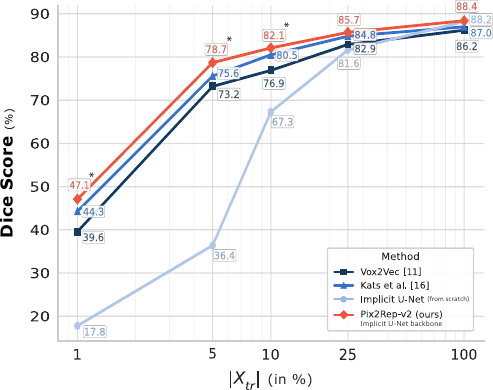}
    \captionof{figure}{\textbf{Multi-organ abdominal CT segmentation results on AMOS}. %Best viewed zoomed-in, in color. We report performance
    Performance (Dice score averaged over the 15 labels) vs.~amount of labeled scans used for fine-tuning $|X_{tr}|$. %(in \%).
    ($\boldsymbol{\ast}$): statistically significant improvement over the best baseline (Wilcoxon signed-rank test, $p < 0.05$).}
    \label{fig: AMOS segmentation results}
\end{minipage}
\end{figure}

\noindent\textbf{Results.} \textbf{Cardiac MRI segmentation} (Fig.~\ref{fig: MRI segmentation results},\ref{fig: Qualitative results}): Results' interpretation is similar across ACDC, M\&Ms, M\&Ms-2. Fine-tuned \texttt{Pix2Rep-v2} (with either backbone) outperforms other methods across all data regimes: \texttt{Pix2Rep-v2} with U-Net is $+9.3$ Dice points above best-of-the-rest \texttt{Pix2Rep} and $+15.0$ Dice points above next-best \texttt{BT-UNet} for $|X_{tr}|=1\%$ on M\&Ms-2. Strikingly, \textit{in-context} \texttt{Pix2Rep-v2} with U-Net performs better for $|X_{S}|=1\%$ than all \textit{fine-tuned} baselines with $|X_{tr}|=1\%$, and $\sim 35$ Dice points above in-context \texttt{ProtoSAM}. It also scales nicely to few-shot, whereas \texttt{ProtoSAM} only natively offers one-shot segmentation. In addition, we get $\times 25$ data-efficiency in few-shot and $\times 5$-$10$ in large data regimes with \texttt{Pix2Rep-v2} pretraining vs.~training from scratch, with identical experimental setups (backbone, pre-processing, training iterations, etc.).

\textbf{Video propagation (cine MRI)} (Fig.~\ref{fig: video propagation results}): \texttt{Pix2Rep-v2} outperforms \texttt{SAM 3} video propagation in zero-shot and almost reaches the performance of \texttt{MedSAM2}, despite \texttt{MedSAM2} having trained on all of ACDC, M\&Ms, M\&Ms-2 scans and GT annotations (including the test data). \texttt{Pix2Rep-v2}'s redundancy reduction-based representations slightly outperform \texttt{Pix2Rep}'s contrastive representations, when coupling them with the propagation mechanism proposed in section~\ref{sec: Methods}.

\textbf{3D abdominal CT segmentation} on AMOS (Fig.~\ref{fig: AMOS segmentation results}): \texttt{Pix2Rep-v2} outperforms other natively 3D self-supervised methods: \texttt{vox2vec}~\cite{goncharov_mikhailand_soboleva_vox2vec_2023} and \texttt{Kats et al.}~\cite{kats_self-supervised_2024}. Furthermore \texttt{Pix2Rep-v2} shows $\times 5$ data-efficiency in low data regimes compared to the implicit U-Net baseline trained from scratch (\textit{i.e.,} it reaches equivalent performance with $\times 5$ fewer annotated scans for fine-tuning).

\section{Discussion and Conclusion}

We presented \texttt{Pix2Rep-v2}, a dense representation learning framework for data-efficient solving of pixel-level tasks, with strong few-shot and in-context capabilities. This opens up new avenues for training next-generation medical imaging foundation models, or for fast development of task- and data-specific AI solutions on premise. Future work will investigate new use cases and tasks (landmark detection, registration), and couple image- with pixel-level representations.

\begin{credits}
\subsubsection{\ackname} This research work is funded by the IP Paris Graduate School, Télécom Paris and the Hi! PARIS interdisciplinary research center. This work was performed using HPC resources from GENCI-IDRIS (Grant 2025-AD011017141).

\subsubsection{\discintname}
The authors have no competing interests to declare that are relevant to the content of this article.
\end{credits}
\bibliographystyle{splncs04}
\bibliography{bibliography}

\end{document}